\documentclass [10pt,twocolumn]{IEEEtran}
\pdfoutput=1
\usepackage{algorithm,algorithmic,amsbsy,amsmath,amssymb,epsfig,bbm,bm,mathrsfs,multirow,amsthm,listings,xcolor}
\usepackage[T1]{fontenc}
\usepackage[latin9]{inputenc}
\usepackage{amsmath}
\usepackage{float}
\usepackage{cite}
\usepackage{algorithm}
\usepackage{array,multirow,graphicx}
\usepackage{color}
\usepackage{makecell}
\usepackage[hidelinks]{hyperref}
\usepackage[flushleft]{threeparttable}

\usepackage{wrapfig}
\colorlet{punct}{red!60!black}
\definecolor{background}{HTML}{F7F7F7}
\definecolor{delim}{RGB}{20,105,176}
\colorlet{numb}{magenta!60!black}

\begin{document}

    \title{CNN-FL for Biotechnology Industry Empowered by Internet-of-BioNano Things and Digital Twins}
    
    \author{Mohammad (Behdad) Jamshidi, \textit{Senior Member, IEEE},  Dinh Thai Hoang, \textit{Senior Member, IEEE}, and\\ Diep N. Nguyen, \textit{Senior Member, IEEE}
\thanks{M. B. Jamshidi, D. T. Hoang, and D. N. Nguyen are with the School of Electrical and Data Engineering, University of Technology Sydney, Australia.}}

    \maketitle


\begin{abstract}
Digital twins (DTs) are revolutionizing the biotechnology industry by enabling sophisticated digital representations of biological assets, microorganisms, drug development processes, and digital health applications. However, digital twinning at micro and nano scales, particularly in modeling complex entities like bacteria, presents significant challenges in terms of requiring advanced Internet of Things (IoT) infrastructure and computing approaches to achieve enhanced accuracy and scalability. In this work, we propose a novel framework that integrates the Internet of Bio-Nano Things (IoBNT) with advanced machine learning techniques, specifically convolutional neural networks (CNN) and federated learning (FL), to effectively tackle the identified challenges. Within our framework, IoBNT devices are deployed to gather image-based biological data across various physical environments, leveraging the strong capabilities of CNNs for robust machine vision and pattern recognition. Subsequently, FL is utilized to aggregate insights from these disparate data sources, creating a refined global model that continually enhances accuracy and predictive reliability, which is crucial for the effective deployment of DTs in biotechnology. The primary contribution is the development of a novel framework that synergistically combines CNN and FL, augmented by the capabilities of the IoBNT. This novel approach is specifically tailored to enhancing DTs in the biotechnology industry. The results showcase enhancements in the reliability and safety of microorganism DTs, while preserving their accuracy. Furthermore, the proposed framework excels in energy efficiency and security, offering a user-friendly and adaptable solution. This broadens its applicability across diverse sectors, including biotechnology and pharmaceutical industries, as well as clinical and hospital settings.
\end{abstract}


%
\IEEEpeerreviewmaketitle

\section{Introduction}
In the biotechnology industry, digital twins (DTs) represent a groundbreaking technology that accurately models humans, organs, microorganisms, and medical systems, along with processes in cyber-physical healthcare systems ~\cite{de2022digital, alazab2022digital}. DTs encompass a wide range of elements, including the Internet of Things (IoT) in healthcare, pharmaceutical industries, clinics, hospitals, and biomedical devices. They oversee the entire lifecycle of biological entities or equipment assets, facilitating monitoring, behavior analysis, control, and adaptation. A fully developed DT transcends a static model or simulation of patients, diseases, organs, or biological assets. It evolves into a dynamic, highly detailed, and nuanced digital counterpart of physical and living entities in these industries, capturing their interactions with the environment. Therefore, DTs represent a seamless fusion of the physical and digital realms, moving beyond the conventional IoT framework of limited inter-connectivity and unidirectional data flow from physical to digital worlds. In the future, industrial DTs will play a vital role in various applications, from drug development to digital healthcare, and in fostering the growth of the Metaverse.\par

Although anticipated to revolutionize the biotechnology industry, employing DTs in biology introduces several significant challenges~\cite{li2021digital}. The first major challenge arises from the inherent complexity and variability of biological systems. Unlike more predictable mechanical systems, biological entities display a high degree of variability and unpredictability, which significantly complicates the process of achieving accurate modeling. Second,  the challenge of data integration presents a significant obstacle. Biological systems often require the integration of various types of data, ranging from molecular to organismal levels, which can be challenging to synchronize and analyze effectively. Third, the necessity of real-time data processing in biology, although crucial, poses a significant challenge. For DTs to be truly effective, they require the capability to process and react to data instantaneously. However, this is particularly challenging due to the intricate and complex nature of biological data. Lastly, ensuring the privacy and security of sensitive biological data is paramount, especially when dealing with human-related data, but it remains a difficult task due to the vast amount of data and its sensitive nature. \par

Recent advancements in the Internet of Bio-Nano Things (IoBNT) are anticipated to address and potentially surmount the prevailing challenges faced by the DTs-enabled biotechnology sector. IoBNT refers to a network of interconnected biological and nanoscale devices capable of communication and data exchange ~\cite{gabrani2023internet}. These devices can include sensors and actuators embedded within biological systems, enabling the real-time monitoring and manipulation of biological processes at a micro or nanoscale. This technology significantly enhances the capability of DTs in the biotechnology industry by providing high-precision, real-time data directly from the biological source. IoBNT devices can monitor and respond to changes at a molecular or cellular level, offering a level of detail and accuracy that greatly improves the modeling and simulation capabilities of DTs. Furthermore, IoBNT can aid in integrating diverse data types by acting as a bridge between the biological system and digital models, ensuring seamless data flow. Additionally, IoBNT can incorporate advanced security measures at the hardware level, offering enhanced data privacy and security, a critical aspect when dealing with sensitive biological data. In summary, the development of IoBNT not only complements but significantly elevates the functionality of DT in the biotechnology industry, addressing key challenges and paving the way for more accurate, real-time, and secure biological system modeling and analysis.\par

However, the integration of IoBNT within DTs-based biology applications presents its own set of distinct challenges, which need to be addressed, particularly in the aspect of sensor allocation in nano scales. Moreover, the presence of numerous nanothings within the same medium can lead to significant interference, resulting in considerable multiuser interference challenges in nanonetworks~\cite{aghababaiyan2022enhanced}. However, integrating IoBNT into the DT-based biotechnology industry presents unique challenges. One of the primary challenges is the processing and interpretation of complex biological data captured by IoBNT devices~\cite{zafar2021systematic}. The data's intricacy and diversity require advanced analytical capabilities to extract meaningful insights. Another challenge is the need for extensive data while ensuring privacy and security, particularly in healthcare settings where data is often sensitive and subject to stringent regulatory protections~\cite{zafar2021systematic}. In this case, Convolutional Neural Networks (CNN) coupled with Federated Learning (FL) offer powerful solutions to these challenges. \par
CNNs excel in handling and analyzing visual and complex data, making them ideal for interpreting the diverse and intricate data generated by IoBNT devices. Their ability to recognize patterns and derive insights from complex biological images allows for more accurate modeling and analysis in the DTs framework~\cite{li2022integrated}. Moreover, CNNs can process the high-dimensional data from IoBNT, providing deeper insights into biological processes and interactions. In the meanwhile, FL addresses the challenge of data availability and privacy. In the highly fragmented landscape of healthcare data, FL enables a collaborative approach to model training without compromising data privacy~\cite{li2022integrated}. By pooling decentralized data insights from various IoBNT sources, FL allows the creation of robust, comprehensive models while maintaining individual data confidentiality. This collaborative training approach enriches CNN's learning, allowing it to access a wider variety of data and improve its accuracy and applicability in diverse biological contexts. Thus, the combination of CNN and FL not only enhances the analytical capabilities in the IoBNT-DT integration but also navigates the critical issues of data privacy and security in the biotechnology industry.\par

In this work, we propose a novel framework aimed at seamlessly integrating IoBNT with DTs technology, specifically tailored for applications in the biotechnology industry. Our approach is multi-faceted and involves several key stages:
\begin{itemize}
    \item Initially, IoTBNTs are deployed to collect a wide array of data from various physical environments. These bio-nano devices are capable of capturing detailed biological and environmental data, which is crucial for accurate modeling and analysis.
    \item The raw data, especially image-based information, is then processed using advanced CNNs. These CNNs are adept at extracting meaningful patterns and insights from the image data, a process that is vital for understanding complex biological interactions and structures.
    \item The trained CNN models are subsequently uploaded to a centralized server. Here, they undergo a federated learning process, wherein multiple models from diverse datasets collaboratively contribute to developing a robust, comprehensive global model. This process ensures the global model benefits from a wide spectrum of data inputs, enhancing its accuracy and generalizability.
    \item After the global model is established, the next step involves creating precise DTs of microorganisms, including bacteria. These DTs are virtual replicas that mimic the behavior, structure, and dynamics of their real-world counterparts, allowing for detailed analysis and simulation in a controlled digital environment.
\end{itemize}
This pioneering framework aims to bridge the gap in integrating IoBNT with DTs for the biotechnology industry by leveraging advanced machine learning techniques, including CNN and FL. By doing so, it addresses a significant challenge in the field, offering a novel approach to data collection, processing, and model creation. We anticipate that this comprehensive framework will not only resolve existing integration challenges but also revolutionize the biotechnology industry. It is expected to pave the way for a new era of highly accurate, efficient, and advanced biotech equipment and communication infrastructure, marking a significant milestone in the evolution of biotechnological applications and research.

\section{DTs for Biotechnology Industry}
DTs are virtual replicas of physical entities or systems that can interact with them and simulate their behavior in real-time. DTs can be used for various purposes, such as designing, optimizing, monitoring, and controlling bioprocesses and bioproducts. DTs can also better understand the complex interactions and dynamics of biological systems, such as cells, tissues, organs, and ecosystems. Some of the typical roles of DTs for biotechnology industry applications are discussed below~\cite{canzoneri2021digital}\cite{herwig2021digital}\cite{gargalo2021towards}.

\subsubsection{Designing and Testing Novel Bioprocesses and Bioproducts}
DTs are designed and tested for bioprocesses and bioproducts in biotechnology and biomanufacturing~\cite{canzoneri2021digital}. This advanced technology utilizes data-driven and model-driven approaches to accurately simulate complex biological systems, significantly enhancing the efficiency of developing vaccines, drugs, and biomaterials. By creating virtual environments, DTs allow for rapid prototyping and process validation, accelerating the development cycle and reducing time-to-market. In the biopharma sector, DTs enable precise modifications in cell fermentation and supply chain management, optimizing operations and fostering innovation~\cite{canzoneri2021digital}. However, the efficacy of DTs hinges on the quality of underlying data and models, and demands substantial computational resources and specialized expertise. Despite these challenges, DTs represent a transformative step forward, promising to expedite the delivery of medical and biotechnological advancements to society. Consequently, a more advanced phase permits the recreation of various scenarios in a simulated environment. This can then be applied to predict future occurrences in real-world settings or to effect changes in the actual events.

\subsubsection{Optimizing Performance, Quality, and Safety}
In bioprocess, DTs are essential, delivering critical insights into the behavior of bioprocesses and the features of bioproducts~\cite{herwig2021digital}. They also contribute to the detection of anomalies, ensuring the stability and consistency of the processes. This involves the real-time examination and fine-tuning of these processes to boost efficiency and augment safety ~\cite{herwig2021digital}. However, it also presents several challenges. Managing the vast amounts of data generated for real-time monitoring and efficiently processing it for actionable insights requires advanced computational capabilities. Accurately modeling complex and dynamic biological systems to predict outcomes and detect anomalies is another significant challenge. Implementing CNN-FL, complemented by IoNBT, offers a promising solution. This integration improves data processing, accuracy, and real-time analysis but introduces complexities in system design and maintenance. Ensuring the accuracy, reliability, and consistent quality and safety standards in these advanced systems, alongside workforce training and adaptation, are critical for harnessing their full potential in optimizing biotechnological processes.

\subsubsection{Monitoring and Predicting Behaviors and Outcomes}
DTs in bioprocess monitoring and prediction showcase significant efficacy. They harness real-time data and advanced modeling techniques to provide deep insights into bioprocess dynamics and bioproduct characteristics, essential for early anomaly detection and correction, thereby ensuring process continuity and product integrity~\cite{gargalo2021towards}. DTs are adept in real-time monitoring, detecting anomalies, and predictive maintenance, playing a key role in ensuring reliability and efficiency in bioprocess operations, marking a substantial step towards improved resilience and optimized process management. Yet, challenges arise in managing large volumes of complex data and accurately modeling dynamic biological systems. Integrating advanced machine learning with IoNBT offers a solution. This integration enhances data processing and analysis, enabling precise, real-time monitoring. Such advancements lead to more accurate predictions, improved anomaly detection, and better process optimization, effectively addressing the challenges and maximizing the potential of DTs in bioprocess monitoring and prediction.

\subsubsection{Holistic View of the Bioprocess Life Cycle}
DTs offer a comprehensive and integrated perspective of the bioprocess life cycle, encompassing research and development, manufacturing, and supply chain management~\cite{gargalo2021towards}. This holistic approach significantly enhances collaboration and streamlines information management across different organizational levels. By providing a detailed digital representation of the entire bioprocess lifecycle, DTs are instrumental in facilitating strategic decision-making, optimizing processes, and effectively managing resources. They consolidate various data sources, fostering cross-functional team collaboration and supporting strategic planning. This integration drives efficiency and sparks innovation throughout the biomanufacturing landscape. The use of DTs in this manner represents a paradigm shift in how bioprocesses are viewed and managed, moving towards a more interconnected and data-driven approach that promises to revolutionize the field~\cite{herwig2021digital}. Implementing DTs across the bioprocess life cycle presents challenges like integrating diverse systems, managing large data volumes, ensuring real-time data accuracy, and fostering cross-functional collaboration. Leveraging IoBNT can enhance precise data collection at a nano-scale, feeding accurate, real-time data into DTs. Concurrently, CNN-FL can efficiently process this complex data, improving analysis and decision-making. IoBNT and CNN-FL address these challenges, enhancing the functionality of DTs in biomanufacturing by ensuring better data quality, advanced processing capabilities, and fostering effective interdisciplinary collaboration, leading to more informed strategic decisions and optimized bioprocess management.

\textit{Summary}: DTs are significantly advancing biomanufacturing, streamlining the design, testing, and optimization of bioprocesses and bioproducts such as vaccines and drugs. Utilizing data-driven and model-driven techniques, DTs aid in swift prototyping, validating processes, and enhancing production efficiency. They improve performance, quality, and safety through artificial intelligence and real-time analytics, enabling predictive maintenance and fault detection. Furthermore, DTs offer a comprehensive perspective on the bioprocess lifecycle, promoting collaboration, informed decision-making, and fostering innovation throughout the industry.

\section{IoBNT and its Potentials for DT-empowered Biotechnology Industry Applications}
\subsection{Internet of Bio-Nano Things (IoBNT): An Overview}
\begin{figure*}[h]
    \centering
    \includegraphics[width=0.75\linewidth]{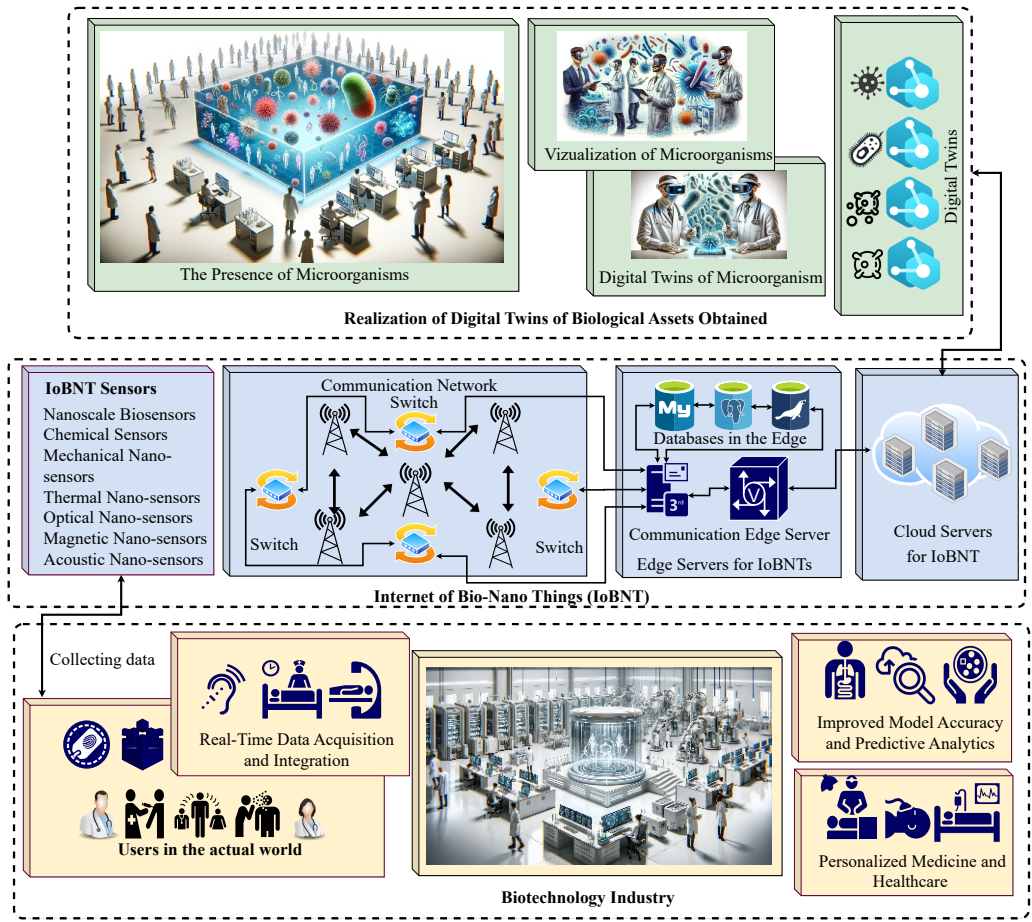}
    \caption{Illustrating how integrating IoBNT can significantly improve applications in the biotechnology industry through DT technology.}
    \label{fig:Fig2}
\end{figure*}

The IoBNT is a visionary concept that merges the realms of nanotechnology, biotechnology, and information technology to forge an interconnected world of tiny biological and nano-engineered devices. Fig. \ref{fig:Fig2} illustrates the potential of IoBNT in enhancing biotechnology industry applications through DT integration. These devices, ranging from molecular sensors and actuators to miniature computational elements, are crafted to interact with biological systems at the cellular or even molecular level. This emerging field endeavors to extend the capabilities of the traditional Internet into the intricacies of biological environments, enabling a network of bio-nano devices that can communicate, process data, and perform specific biological functions. The development of IoBNT is fueled by breakthroughs in nano-fabrication techniques, molecular biology, and wireless communication, which collectively enable the design and operation of devices that can function reliably within living organisms or in direct contact with biological materials.
The significance of IoBNT for the future of the biotechnology industry is profound.\par

In healthcare, IoBNT technologies are set to revolutionize medical diagnostics and treatment, offering new avenues for personalized medicine. By providing tools for targeted drug delivery and real-time monitoring of physiological processes, IoBNT enables treatments that are precisely tailored to individual patient needs, thereby increasing efficacy and minimizing side effects. This precision medicine approach heralds a new era in healthcare where treatments are adapted to the unique genetic and molecular profile of each patient. Furthermore, in the field of environmental biotechnology, IoBNT devices offer unparalleled sensitivity and specificity in detecting environmental pollutants and pathogens, leading to more effective monitoring and management of ecological systems. The ability of IoBNT to gather and analyze data at an unprecedented scale and depth holds the key to unlocking new insights in biological research, accelerating the pace of discovery and innovation in biotechnology. The integration of IoBNT into the biotechnology industry is not just an enhancement of existing technologies; it represents a paradigm shift towards more efficient, precise, and personalized biotechnological applications, fundamentally changing how we interact with and manipulate biological systems for the betterment of health, environment, and society~\cite{ kuscu2021internet}.
\subsection{Potential Applications of IoBNT for DT-Enabled Biotechnology Industry} 
The IoBNT is transformative in enhancing DTs enabled biotechnology industry applications. By integrating IoBNT with DTs, the biotechnology industry can achieve unprecedented levels of precision, efficiency, and innovation. The potential of IoBNT in this field is immense, opening up new avenues for research and application. This integration promises to bring about a new era in biotechnology, with smarter and more efficient systems, leading to significant advancements in various fields such as medicine, environmental monitoring, and agricultural biotechnology.
\subsubsection{Real-Time Data Acquisition and Integration} 
IoBNT can provide real-time data on biological systems and their dynamics, such as cellular activities, molecular interactions, metabolic pathways, gene expressions, etc. This data is crucial for understanding complex biological processes in real-time, leading to more accurate and efficient biotechnological applications~\cite{ kuscu2021internet}. The integration of this data with other information sources, like medical records, environmental factors, or user preferences, creates a more holistic and detailed understanding of biological systems. This comprehensive view is essential for developing more targeted and effective biotechnological solutions, enhancing the capacity to address complex biological challenges.
\subsubsection{Improved Model Accuracy and Predictive Analytics} IoBNT can enable continuous and fine-grained calibration and validation of the DT models, which can improve their accuracy and reliability ~\cite{zafar2021systematic}. The ability to update and refine models in real-time based on new data ensures that the DTs remain relevant and accurate, reflecting the current state of the biological systems they represent. IoBNT can also provide feedback and feedforward mechanisms to adjust the models according to the changes in the system or the environment. This adaptability enhances the predictive analytics capabilities of the DTs, leading to better forecasting, optimization, and decision-making in various biotechnological applications. 
 \subsubsection{Personalized Medicine and Healthcare} IoBNT can enable personalized and precision medicine and healthcare, where the treatments and interventions are tailored to the specific needs and characteristics of each individual~\cite{akyildiz2015internet}. By monitoring individual health status and responses to therapy, IoBNT can provide critical data for personalized healthcare. DTs can then simulate the patient's physiology and the effects of different drugs or devices, leading to more effective and safer treatments. This approach could significantly improve patient outcomes and satisfaction, marking a major advancement in the field of medicine and healthcare. Personalized healthcare solutions, enabled by the integration of IoBNT and DTs, represent a future where treatments are not only more effective but also safer and more attuned to individual patient needs. 

\textit{Summary}: The section discusses the IoBNT, a concept combining nanotechnology, biotechnology, and information technology to create interconnected bio-nano devices. These devices integrate with biological systems, enhancing medical diagnostics, treatments, and environmental biotechnology. IoBNT's integration with DTs in the biotechnology industry promises significant advancements in precision, efficiency, and innovation, particularly in personalized medicine and healthcare. This integration leads to more effective, tailored treatments and a profound impact on healthcare and environmental management.

\section{FNN-CNN for Digital Twin integrated with IoBNT for Biology Applications }
This section introduces a groundbreaking framework specifically formulated to tackle the inherent challenges associated with the implementation of the IoBNT and DTs technology in the biotechnology sector. Addressing these intricate complexities necessitates a strategic and innovative approach. Consequently, we propose a synergistic integration of CNNs, FL, and IoBNT. This integrative methodology is designed to leverage the distinctive capabilities of each element, thereby enhancing the overall efficacy, precision, and functionality of IoBNT and digital twin applications in these advanced fields. The confluence of these technologies promises to revolutionize data processing and analysis, offering a more nuanced and effective deployment of bio-nano technologies. This, in turn, is anticipated to yield substantial advancements in biotechnological methodologies and digital health interventions. 

\begin{figure}[t]
    \centering
    \includegraphics[width=.95\linewidth]{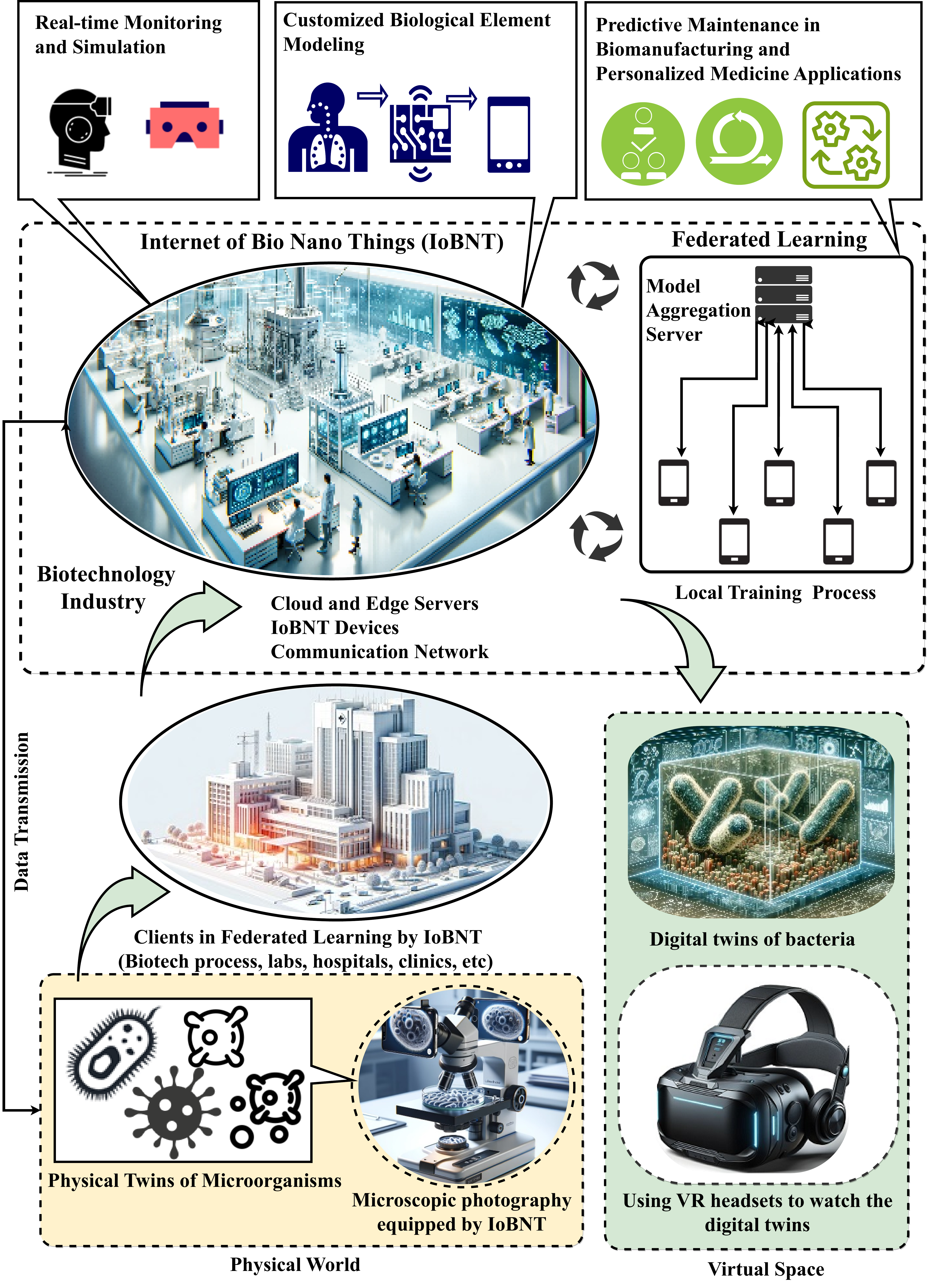}
    \caption{The convergence of IoBNT and FL for advanced DTs in biotechnology: micro-level data analysis with ML-driven techniques.}
    \label{fig:Fig1}
\end{figure}

Fig.~\ref{fig:Fig1} illustrates the synergistic integration of IoBNT with FL in enhancing DTs in the biotechnology sector. This fusion combines the advanced capabilities of IoBNT with the machine learning-driven approaches of FL to facilitate precise analysis of micro-level biological and environmental data. The figure highlights the potential of this integration in the biotechnology industry, emphasizing its role in creating a comprehensive digital environment. This environment is pivotal for various applications including real-time monitoring and simulation, customized biological element modeling, and predictive maintenance in biomanufacturing, as well as personalized medicine. FL contributes to the process by enabling various clients, such as hospitals and laboratories, to collaborate in the pattern recognition of microorganisms across different locations. This collaborative approach not only enhances efficiency but also minimizes energy consumption. In essence, the process begins with the IoBNT nodes capturing images of the physical twins of microorganisms. These images are then processed across different sites, contributing to an updated global model through FL. Subsequently, the refined and accurate digital twins of these microorganisms are constructed in a virtual space.

\subsection{Proposed System Model}
The proposed system model is designed for a broad spectrum of biotechnology applications, focusing primarily on the digitalization and virtualization of microorganisms from various physical environments. This model includes the physical environments like the biotech process, where the presence and categorization of different bacteria types, both beneficial and harmful, are crucial for health-related applications, including food or drug development and fermentation processes.

\begin{figure*}[h]
    \centering
    \includegraphics[width=0.9\linewidth]{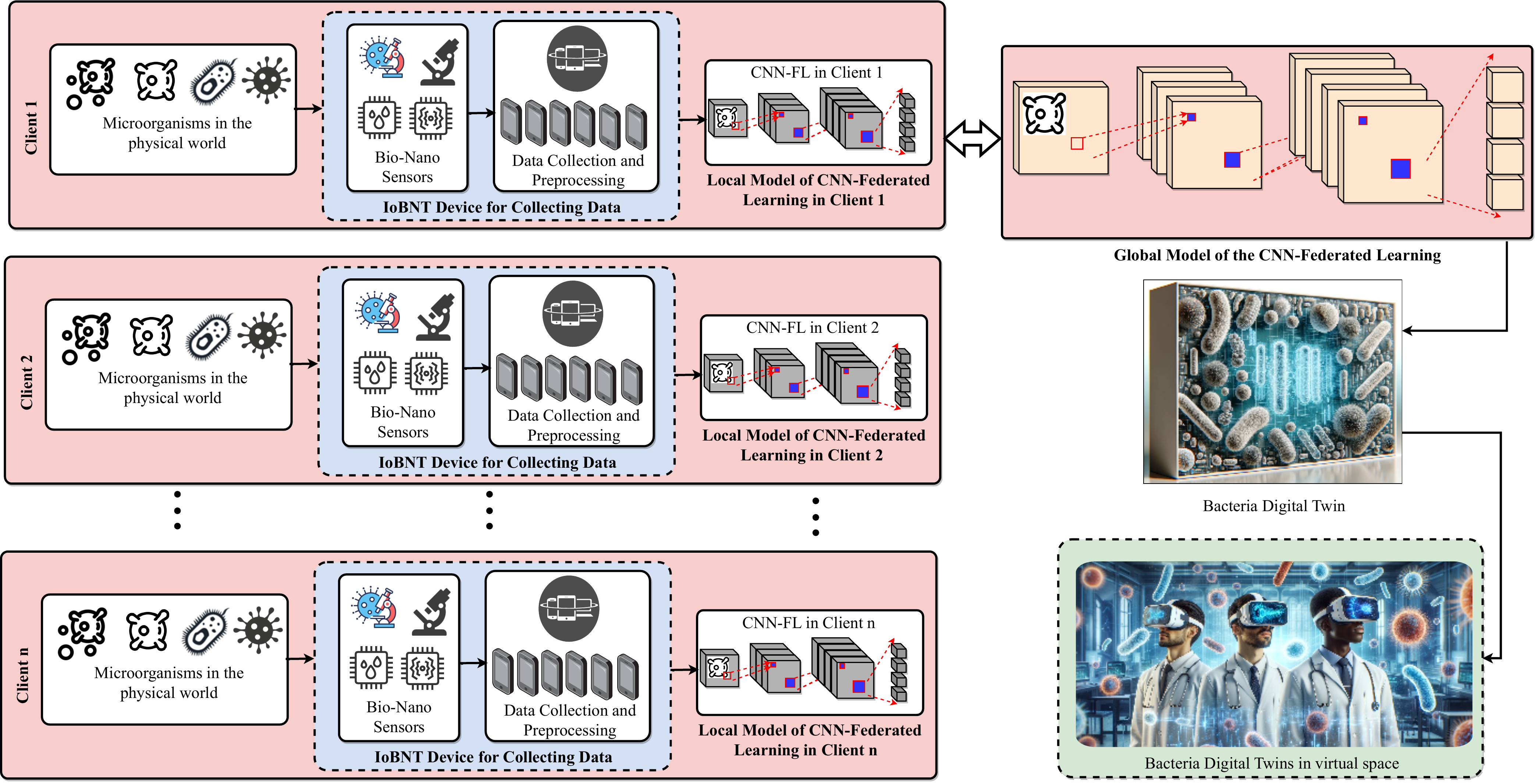}
    \caption{The framework for creating bacteria DTs, utilizing CNN and FL integrated with IoBNT technology. This framework features a cost-effective approach for microscopic photography, compatible with wireless communication protocols. The system encompasses various clients including clinics and hospitals, contributing to a global model that supports the seamless realization of a sustainable, interactive DT environment.}
    \label{fig:Fig3}
\end{figure*}

Fig.~\ref{fig:Fig3} illustrates the proposed system model including the proposed solution. A key component of the system is the concept of "physical twins" of bacteria. These are essentially real-world microbial entities that are present in various environments. The first major challenge addressed by the system is the accurate recognition and digitalization of these microorganisms. This involves categorizing them into their respective groups based on specific characteristics and behaviors. This step is crucial as it lays the foundation for their subsequent virtual representation and analysis. The second challenge revolves around establishing a specific communication method tailored to such environments.\par
In biotechnology applications, where precision and accuracy are paramount, the choice of a communication network is vital. It needs to be robust enough to handle the specificities of microbial data, which can be complex and multifaceted. This network forms the backbone of the system, enabling the seamless transmission and reception of data between the physical and virtual domains. The third issue that the system model aims to tackle is ensuring an efficient way of processing data. This is not limited to security aspects but extends to energy considerations as well. In an era where energy efficiency is increasingly prioritized, it is crucial for systems to not only ensure data security but also to operate in a manner that conserves energy. This is especially pertinent in settings like clinics or research facilities, where large volumes of data are handled regularly. Finally, the model integrates these components into a virtual platform~\cite{hoang2023metaverse}. This platform serves as a virtual space where the digital versions of the microorganisms, transmitted from various physical locations, are represented. The virtual platform offers an innovative and interactive medium for scientists and researchers to analyze, experiment with, and understand these microorganisms in a comprehensive and immersive virtual environment. This integration of physical microbial data into a virtual space opens up new possibilities for research and development in biotechnology, enhancing both the depth and scope of analyses and applications.\par

\subsection{Proposed Solution}
In this section, we delve into the deployment of FL in our proposed framework, emphasizing its integration with the IoBNT to enhance DTs applications in biotechnology. FL emerges as a crucial element in tackling scalability issues prevalent in IoT and DT systems. Its decentralized data processing capability allows IoBNT devices to collaboratively learn from data without sharing the raw data, thus preserving privacy and reducing bandwidth demands. This decentralized approach is especially beneficial for DTs, facilitating the integration and analysis of extensive, distributed datasets from IoBNT devices, and thereby improving the accuracy and responsiveness of digital replicas. The IoBNT framework plays an essential role in processing and analyzing biological data at a very detailed level, enabling precise and efficient healthcare solutions. The challenge lies in integrating FL within the IoBNT framework, but its successful implementation is expected to revolutionize data processing in biotechnology, merging decentralized learning with nano-scale precision. Furthermore, the framework incorporates CNNs for creating DTs for microorganisms and other objects. CNNs, known for their effectiveness in processing and analyzing complex visual data, are ideally suited for creating accurate digital representations of physical entities, a critical aspect for DTs. Integrating FL with CNNs significantly enhances this process in IoT environments. It enables the system to handle vast amounts of data from various sources without centralizing data storage. This synergy between CNN, FL, and IoBNT effectively addresses key challenges in biotechnology, such as privacy, bandwidth, accuracy, and scalability, leading to more advanced and efficient digital twinning processes and paving the way for more sophisticated and energy-efficient DT applications.

As illustrated in Fig.~\ref{fig:Fig3}, the FL firstly emerges as a pivotal component in addressing scalability issues commonly found in IoT systems, including DTs. The FL's decentralized data processing capability allows IoBNT devices to collaboratively learn from data without the need to share the raw data itself. This not only preserves the privacy of sensitive biological data but also significantly reduces bandwidth demands. In the context of DTs, this decentralized approach enables the integration and analysis of extensive, distributed datasets from IoBNT devices, thereby enhancing the accuracy and responsiveness of the digital replicas. The IoBNT framework is crucial for processing and analyzing biological data. It provides a robust network of bio-nano devices that can interact with biological systems at a very granular level, enabling precise and efficient healthcare solutions. The integration of FL within IoBNT is a key challenge, but once implemented, it promises to revolutionize data processing in biotechnology by bringing together the best of decentralized learning and nano-scale precision.\par

\subsection{Simulation Results}
This work studies two datasets to evaluate the efficiency of the proposed approach. The primary dataset, Digital Images of Bacteria Species (DIBaS)~\cite{zielinski2017deep}, initially contained 689 RGB images of 33 bacteria species. Recognizing the need for more data, additional images, which were extracted by a specific IoBNT, were added to the initial dataset, bringing the total to 2722 RGB images. The designed IoBNT for this purpose includes a Nikon E200 microscope at 100x magnification equipped with a smartphone for micrography. After categorization of the new images, we systematically added each image to the appropriate category in DIBaS, ensuring the dataset remained organized and accurate. The bacteria samples are sourced from blood, urine, and skin. The experimental setup comprises two approaches, i.e., an FL framework utilizing a CNN across three clients, and a centralized CNN approach. The CNN methodology incorporates MobileNetV2~\cite{sandler2018mobilenetv2}, a widely recognized architecture for its efficiency in mobile and embedded vision applications. The FL setup is powered by Flower (flower.dev), a versatile FL framework. Flower operates on a server-client architecture, where a central server orchestrates the training process across multiple clients. It manages the synchronization of training rounds and aggregates the model updates received from each client, ensuring efficient and collaborative learning across the distributed network.\par
The setup involves configuring CNNs for precise pattern recognition and image processing, essential for accurately modeling bacterial images in DTs. Simultaneously, the FL is implemented to facilitate decentralized learning across IoBNT devices, each equipped with CNN capabilities. This enabled data processing directly on each device, enhancing privacy and reducing data transfer. During simulations, data from multiple IoBNT devices was collected and used to train the CNN models, enhancing their accuracy in recognizing bacterial behaviors. The integration of CNN-FL with IoBNT proved critical, combining nano-scale biological data analysis with robust, decentralized data processing. The results showed a marked improvement in the accuracy of bacterial behavior recognition compared to traditional methods. This advancement underscores the potential of CNN-FL in revolutionizing biotechnological applications, particularly in improving healthcare solutions and understanding complex biological systems. 

In our analysis, we employ the confusion matrix due to its comprehensive evaluation capabilities. The confusion matrix is one of the most important metrics for evaluating machine vision models, offering a detailed representation of performance by distinguishing between correct predictions and different types of errors. This distinction is critical in machine vision, where the consequences of misclassifications can vary greatly. It consists of four elements, i.e., True Positives (TP), True Negatives (TN), False Positives (FP, or Type I errors), and False Negatives (FN, or Type II errors). This matrix is particularly useful in situations with imbalanced classes or varying error costs. It allows for the calculation of accuracy, precision, recall, and F1-score, providing a holistic view of the classifier's effectiveness in machine vision tasks. In conclusion, the confusion matrix is essential for evaluating the performance of classification models, offering a comprehensive view of their ability to differentiate between classes.

\begin{figure}[h]
    \centering
    \includegraphics[width=.90\linewidth]{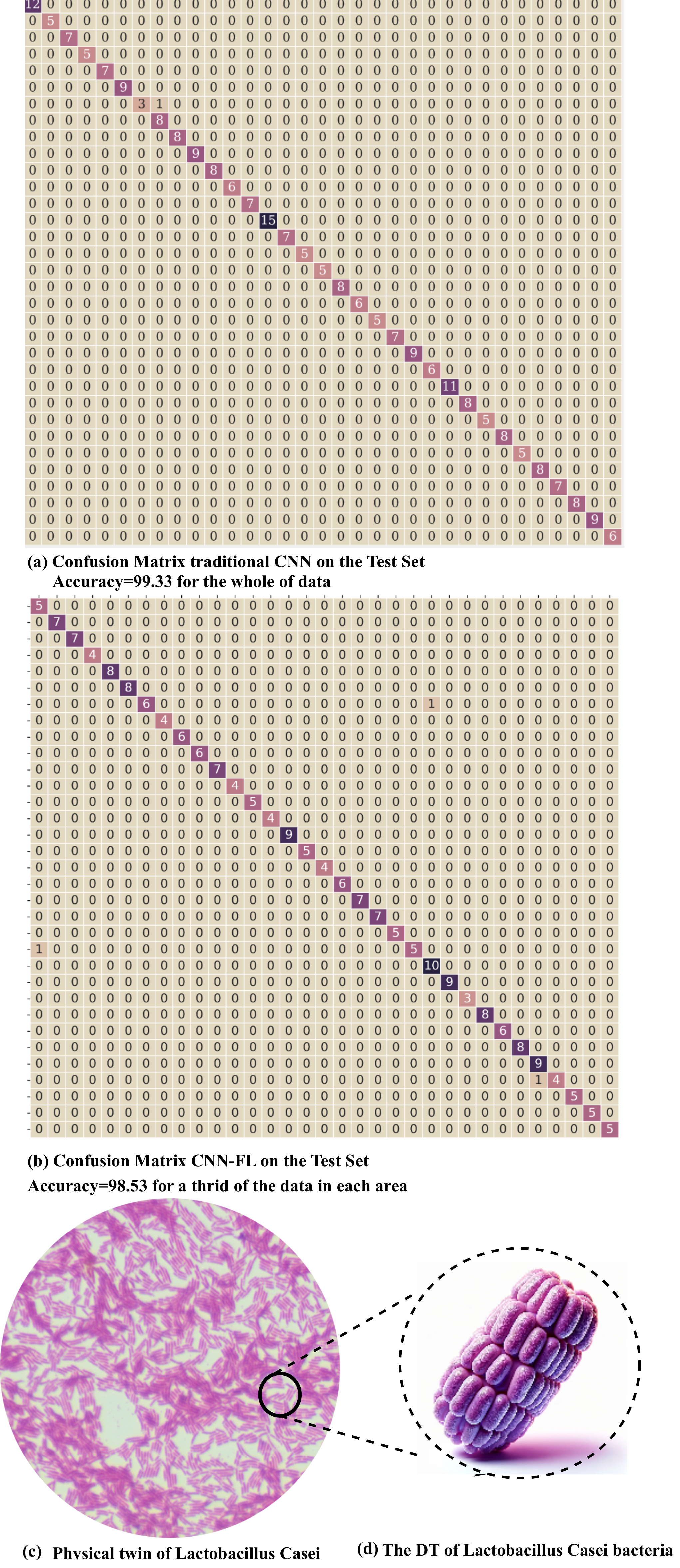}
    \caption{The evaluation and visualization of the proposed approach: (a) The confusion matrix of the conventional CNN trained by the whole of the dataset. (b) The confusion matrix of the proposed method for a third part of data in each area. (c) A detailed physical representation of the bacteria under study, highlighting specific morphological characteristics relevant to the IoBNT application. (d) The DT of the selected group of bacteria indicated by the black circle in (c).}
    \label{fig:Fig4}
\end{figure}

Fig.~\ref{fig:Fig4} illustrates the performance of the proposed approach. Specifically, Fig.~\ref{fig:Fig4}(a)  shows the confusion matrix for the centralized CNN trained and tested by data extracted through IoBNT. It is important to note that in this method, we assume that there is one centralized node which is able to collect and train all the data in the network. As a result, the results obtained by this approach usually achieve the highest performance, and thus, they can be served as the upper bounds to compare with our proposed approach using FL. Fig.~\ref{fig:Fig4}(b) presents the confusion matrix for the proposed CNN-FL model framework using the same data. The results obtained in Fig. 4(b) clearly show the efficiency of the proposed approach as they are very close to those of the upper bound approach (i.e., the centralized solution). Alternatively, Fig.~\ref{fig:Fig4}(c) provides a detailed visualization of the physical representation of the Lactobacillus Casei bacteria extracted by the designed IoBNT. This is one of the 2722 RGB images used in this work. It highlights the morphological characteristics and nuances of the bacteria, which are crucial for understanding their behavior and interaction within the proposed framework. Fig.~\ref{fig:Fig4}(d) illustrates the DT of a specific group of Lactobacillus Casei bacteria, which are highlighted by a black circle in Fig.~\ref{fig:Fig4}(c). To create this digital counterpart, we employed an advanced generative AI tool. The resulting DT accurately replicates the features of the selected bacteria group shown in Fig.~\ref{fig:Fig4}(c). While the black circle highlighted in Fig.~\ref{fig:Fig4}(c) depicts the physical twin of a group of real bacteria within a container, Fig.~\ref{fig:Fig4}(d) provides a detailed digital representation of this specific subset.

\section{Open Issues and Future Research Directions}
Integrating the Internet of BioNano Things (IoBNT) into the DTs-based biotechnology industry is an exciting yet challenging area of research, with several open issues and potential directions for future exploration.

\subsubsection{Design and Fabrication of Bio-Nano Things (BNTs)}
BNTs are crucial for IoBNT, requiring advanced design and fabrication using synthetic biology and nanotechnology. These BNTs must be capable of sensing, processing, communicating, and actuating within biological domains, and interfacing with cyber domains. Key aspects to be addressed include ensuring biocompatibility, biodegradability, and energy efficiency. Future research could focus on developing new materials and techniques for BNT fabrication, exploring sustainable energy sources for BNTs, and enhancing the interface between BNTs and biological systems to improve performance and reliability.

\subsubsection{Communication and Networking Methods for IoBNT}
IoBNT's reliance on unconventional communication methods like molecular communication introduces challenges such as noise, interference, and molecular coding. Networking methods specific to IoBNT, addressing routing, synchronization, and security within the bio-nano domain, are crucial. Research directions could include the development of more robust molecular communication techniques, innovative networking protocols tailored for bio-nano environments, and security frameworks that protect against potential cyber threats in these unique settings.

\subsubsection{Bio/Cyber and Nano/Macro Interfaces}
For IoBNT, interfacing with the cyber domain of the Internet and the macro domain of the biotechnology industry is vital. Developing bio/cyber interfaces (like biosensors and bioactuators) and nano/macro interfaces (such as nanofluidic channels and microfluidic chips) is essential. Future research could explore advanced materials and technologies for interface development, methods to enhance signal translation and data integration between these domains, and innovative design principles for creating more efficient and scalable interface systems.

\subsubsection{Digital Twin Modeling and Simulation}
DT models need to accurately represent biological systems and incorporate their dynamics and interactions. These models should be reliable and scalable. Developing simulation tools and methods to analyze, optimize, and predict the behavior and performance of both the biological systems and the IoBNT is a critical research area. Potential directions include enhancing model accuracy, incorporating advanced data analytics and machine learning techniques for better predictions, and developing scalable simulation platforms that can handle complex, multi-scale biological systems.

Overall, each of these areas offers substantial opportunities for innovation and advancement in the field of IoBNT and its integration into the biotechnology industry, paving the way for breakthroughs in personalized medicine, environmental monitoring, and beyond.

\section{Conclusions}
This work highlights the significant role of digital twins (DTs) in enhancing the biotechnology industry, focusing on digital models of biological elements, and the required more efficient types of Internet of Things (IoT) infrastructures in this area. It addresses the challenges in applying DT technology at micro and nano scales, particularly in modeling complex entities like bacteria. The introduced framework innovatively merges the Internet of Bio-Nano Things (IoBNT) with advanced machine learning methods such as convolutional neural networks (CNN) and federated learning (FL), effectively improving the efficiency, reliability, and dependability of DTs in the biotechnology industry. A key contribution of this research is the development of this novel framework, enhancing the reliability and safety of microorganism DTs while maintaining accuracy. The approach also stands out in energy efficiency and security, providing a user-friendly, flexible platform. Its versatility broadens applicability to various sectors, including pharmaceuticals, healthcare systems, and biomanufacturing, showcasing its extensive potential impact. Future research will explore extending this framework to personalized medicine and environmental monitoring, leveraging its adaptability for broader scientific and societal benefits.\par

\ifCLASSOPTIONcaptionsoff
  \newpage
\fi

\bibliographystyle{IEEEtran}
\bibliography{CNNFLBacteriaMagezine.bib}

\end{document}